\documentclass[conference]{IEEEtran}
\usepackage{times}

\usepackage[numbers]{natbib}
\usepackage{multicol}
\usepackage[bookmarks=true]{hyperref}

\let\labelindent\relax
\usepackage{enumitem}
\usepackage{graphicx}
\usepackage{booktabs}
\usepackage{multirow}
\usepackage{colortbl}
\usepackage[table]{xcolor}
\usepackage{subcaption}
\usepackage{stfloats}
\usepackage{array}
\usepackage{amsmath}
\usepackage{algorithm}
\usepackage{algpseudocode}
\usepackage{amsfonts}

\usepackage[table]{xcolor} 
\usepackage{booktabs}
\usepackage{multirow}

\definecolor{gaincolor}{RGB}{34, 139, 34} 
\definecolor{losscolor}{RGB}{178, 34, 34} 
\definecolor{highlightrow}{gray}{0.92}    

\newcommand{\cg}{\cellcolor{highlightrow}} 
\newcommand{\gain}[1]{\textcolor{gaincolor}{\scriptsize\textbf{(+#1)}}}
\newcommand{\loss}[1]{\textcolor{losscolor}{\scriptsize(#1)}}

\definecolor{gray10}{gray}{0.95}

\begin{document}

\title{Rethinking Visual-Language-Action Model Scaling: Alignment, Mixture, and Regularization}





\author{Ye Wang$^{12}$\authorrefmark{1} \quad Sipeng Zheng$^{2}$\authorrefmark{1} \quad Hao Luo$^{23}$\authorrefmark{1} \quad Wanpeng Zhang$^{23}$\authorrefmark{1} \quad Haoqi Yuan$^{23}$ \quad Chaoyi Xu$^{2}$ \\ \vspace{1mm} Haiweng Xu$^{23}$ \quad Yicheng Feng$^{23}$ \quad Mingyang Yu$^{1}$ \quad Zhiyu Kang$^{1}$ \quad Zongqing Lu$^{23}$ \quad Qin Jin$^{1}$\authorrefmark{2} \\
$^1$Renmin University of China \quad$^2$BeingBeyond \quad$^3$Peking University\\
\authorrefmark{1}Equal contribution\quad\authorrefmark{2}Corresponding author \vspace{-3mm}}

\maketitle

\begin{abstract}
While Vision–Language–Action (VLA) models show strong promise for generalist robot control, it remains unclear whether---and under what conditions---the standard ``scale data" recipe translates to robotics, where training data is inherently heterogeneous across embodiments, sensors, and action spaces. 
We present a systematic, controlled study of VLA scaling that revisits core training choices for pretraining across diverse robots.
Using a representative VLA framework that combines a vision–language backbone with flow-matching, we ablate key design decisions under matched conditions and evaluate in extensive simulation and real-robot experiments.
To improve the reliability of real-world results, we introduce a Grouped Blind Ensemble protocol that blinds operators to model identity and separates policy execution from outcome judgment, reducing experimenter bias.
Our analysis targets three dimensions of VLA scaling. 
(1) Physical alignment: we show that a unified end-effector (EEF)-relative action representation is critical for robust cross-embodiment transfer. 
(2) Embodiment mixture: we find that naively pooling heterogeneous robot datasets often induces negative transfer rather than gains, underscoring the fragility of indiscriminate data scaling. 
(3) Training regularization: we observe that intuitive strategies, such as sensory dropout and multi-stage fine-tuning, do not consistently improve performance at scale. 
Together, this study challenge some common assumptions about embodied scaling and provide practical guidance for training large-scale VLA policies from diverse robotic data.
Project website: \url{https://research.beingbeyond.com/rethink_vla}
\end{abstract}

\IEEEpeerreviewmaketitle

\section{Introduction}

Vision–Language–Action (VLA) models~\cite{brohan2022rt, intelligence2025pi05} have become a promising direction for general-purpose embodied AI.
Following the scaling trends of vision–language models (VLMs)~\cite{comanici2025gemini2.5, qwen3-vl}, robotics is moving from single-task policies to generalist policies that aim to solve many tasks across different environments. 
This shift is supported by recent large robotic datasets~\cite{o2024openx, bu2025agibot} with thousands of hours from diverse robot platforms, collected in both simulation and the real world.

A common belief is that, as in language modeling, improved generalization in robotics will primarily emerge from scaling data and model size. 
However, scaling in robotics introduces physical and system-level challenges that do not arise in text or image generation.
Robotic data is inherently heterogeneous: 
robots differ in kinematics, joint limits, control frequencies, sensing modalities, and action spaces, which are often incompatible by design.  As a result, it remains unclear whether scaling heterogeneous robotic data reliably leads to positive transfer, or whether embodiment differences introduce interference that limits performance. 

These challenges raise several open questions that are central to scaling VLA models: \emph{What action representations best align heterogeneous embodiments?} \emph{When does mixing data across robots help, and when does it hurt?} And \emph{how effective are commonly used regularization strategies when applied at scale in embodied settings?}

In this work, we address these questions through a controlled empirical study of VLA scaling. 
Instead of proposing a new architecture, we build a controlled testbed based on a representative VLA framework that combines a VLM backbone with flow matching~\cite{liu2022flow, black2024pi_0}. 
We treat the training pipeline itself as the primary object of study and systematically ablate key design choices under matched conditions. 
Our study is organized around three pillars:
\textbf{(1) Physical Alignment.} 
We examine how coordinate frames and action representations affect general control performance across different setups.
\textbf{(2) Embodiment Mixture.} 
We study how performance changes with different mixtures of heterogeneous data.
Specifically, we investigate whether pooling trajectories across different robots promotes positive transfer, or if cross-embodiment variation can introduce interference.
\textbf{(3) Training Regularization.} 
We evaluate the scalability of training modifications, such as sensory dropout and curriculum learning, to determine if these approaches provide tangible benefits in a large-scale pretraining regime.

Reliable evaluation is also critical for drawing conclusions about scaling.
Real-world testing can be influenced by experimenter bias, including operator familiarity with a policy and small adjustments during execution.
To ensure reliable evaluation, we propose a Grouped Blind Ensemble protocol for real-world experiments: operators execute tasks without knowing which model variant is being tested, and we separate model inference from outcome judgment.
By decoupling policy execution from outcome judgment and blinding operators to model identity, this protocol reduces human bias and improves the objectivity of real-robot evaluation.

Our main findings provide practical guidance for training large-scale VLA models: 
1) End-effector (EEF) relative action space consistently outperforms other representations, establishing itself as a reliable default for VLA training;
2) Scaling cross-embodiment data is challenging because indiscriminately mixing heterogeneous robotic data frequently degrades performance, highlighting the fragility of naive data scaling;
3) Intuitive training modifications, such as randomized sensory dropouts and multi-stage curricula, do not reliably translate into gains at scale;
4) Our proposed grouped blind ensemble protocol for real-world evaluation significantly reduces human bias in measuring robotic performance.

\begin{figure*}[!ht]
\centering
\includegraphics[width=\textwidth]{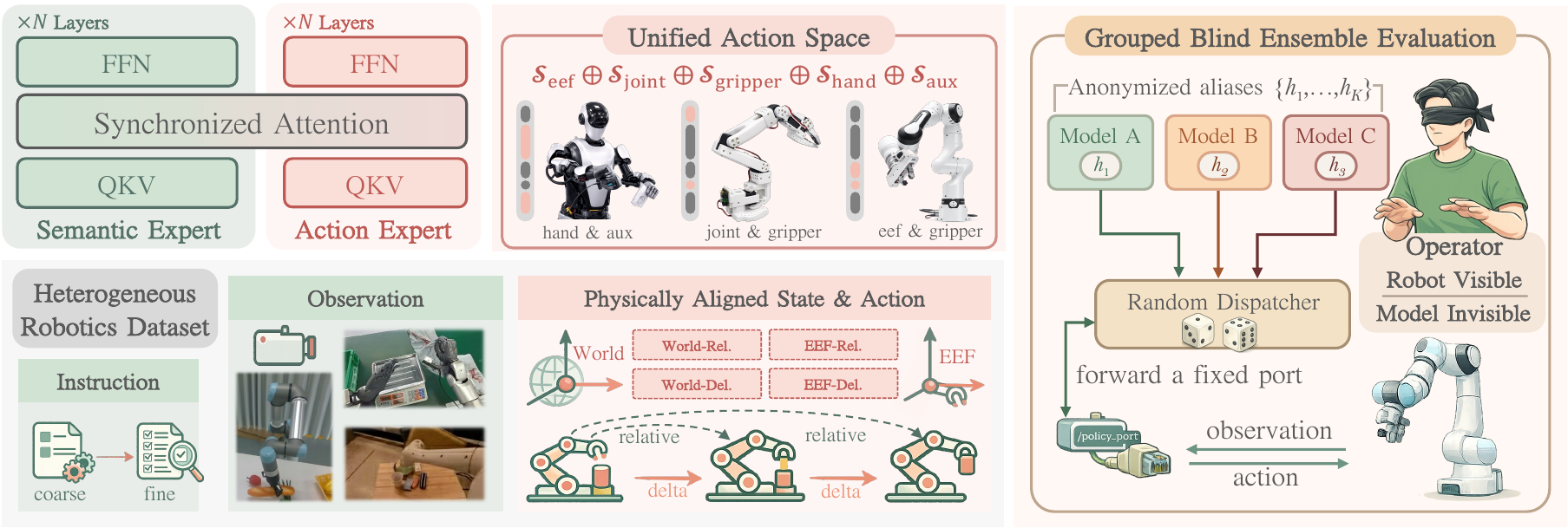}
\caption{
Overview of our systematic VLA analysis framework, comprising the Mixture-of-Transformers architecture, physically aligned action spaces, and the Grouped Blind Ensemble protocol.
}
\label{fig:method_eval}
\vspace{-1em}
\end{figure*}

\section{Related Work}
\noindent\textbf{Vision-Language-Action Models.}
Research in robotic manipulation has evolved significantly, transitioning from narrowly defined, single-task policies to more versatile systems that are trained on extensive and varied datasets~\cite{berscheid2019robot, dasari2019robonet, fang2023rh20t, shafiullah2023bringing}.
Vision–Language–Action (VLA) models~\cite{brohan2022rt,zitkovich2023rt,kim2024openvla,janner2022planning,chi2025diffusion} exemplify this trend by transferring the perceptual and reasoning priors of pre-trained vision-language models (VLMs)~\cite{steiner2024paligemma, li2025eagle, wang2024qwen2vl, zhang2025bpe, zhang2025unified, hao2025openmmego, feng2025videoorion} into embodied control through fine-tuning on robot demonstrations.
Although many VLAs share similar backbone architectures, their action generation choices vary widely.
Early works often cast continuous control as discrete tokens and decode actions autoregressively to better match VLM-style objectives~\cite{kim2024openvla, qu2025spatialvla}; this design eases transfer but can increase inference cost and struggle with high-precision, high-DoF execution.
In contrast, more recent approaches adopt diffusion-style objectives~\cite{ho2020denoising,li2024cogact, janner2022planning, liu2024rdt, liang2025discrete} or flow-matching formulations~\cite{lipman2022flow,nvidia2025gr00t,black2024pi_0,intelligence2025pi05,luo2026being05,cai2026internvla} to generate robot action chunks.
Despite these advances, it remains challenging to consistently translate high-level VLM priors into fine-grained control across diverse embodiments.
In contrast to proposing a new architecture, our work focuses on how training choices such as action parameterization, cross-embodiment data mixture, and regularization affect VLA scaling under controlled conditions.

\noindent\textbf{Robotic Data for Pre-Training.}
The efficacy of Vision-Language-Action (VLA) training is closely linked to the scale and diversity of the robotic data utilized.
Numerous initiatives, such as Open X-Embodiment~\cite{o2024openx}, Droid~\cite{khazatsky2024droid} and BridgeData~\cite{walke2023bridgedata}, compile significant demonstration hours across various tasks, environments, and platforms.
Community-driven datasets and benchmarking efforts further enhance accessibility and coverage in this domain~\cite{shukor2025smolvla}.
To better capture fine-grained manipulation, recent datasets are focusing on bimanual coordination, dexterous actions, and more sophisticated interaction scenarios.
Notable examples include AgiBot World~\cite{bu2025agibot}, and RoboCOIN~\cite{wu2025robocoin} and RoboMIND~\cite{wu2024robomind, hou2025robomind2}, while Open Galaxea~\cite{jiang2025galaxea} is dedicated to mobile manipulation tasks. 
Nevertheless, much teleoperated data is still collected in constrained settings.
Synthetic data~\cite{ye2025dex1b, chen2025internvla-m1} can alleviate data scarcity, but the sim-to-real gap remains a persistent obstacle. 
More fundamentally, these datasets are heterogeneous by construction, spanning different kinematic structures, sensing/control conventions, and action spaces, which complicates naive pooling and transfer.
Our study directly examines this heterogeneity and provides controlled evidence on when mixing data across embodiments helps or hinders, together with a bias-reduced real-robot evaluation protocol.

\section{Methodology and Evaluation Protocol}

To study how vision–language–action (VLA) models scale under heterogeneous training data, we build (i) a controlled testbed for generalist architectures and (ii) a rigorous physical evaluation protocol. 
In this section, we describe the model, the cross-embodiment unification used for physical evaluation, and a bias-resistant testing procedure, as illustrated in Figure \ref{fig:method_eval}.

\subsection{Mixture-of-Transformers with Flow-Matching Control}

Our goal is to combine high-frequency, precise control with strong semantic reasoning. 
Following prior work, we adopt a Mixture-of-Transformers (MoT) design.~\cite{deng2025bagel,black2024pi_0}.

\noindent\textbf{Dual Experts with Layerwise Shared Attention.}
The model uses two transformer backbones in parallel.
The \textbf{Semantic Expert} ($\mathcal{E}_{\text{sem}}$) is initialized from a pretrained VLM to preserve visual and language priors.
The \textbf{Action Expert} ($\mathcal{E}_{\text{act}}$) is a randomly initialized and trained for control.
For efficiency, it uses a smaller hidden dimension, but it matches the Semantic Expert in depth.
We split the input sequence into four token groups: $\mathbf{X} = \{\mathbf{X}_V, \mathbf{X}_L, \mathbf{X}_S, \mathbf{X}_A\}$, corresponding to visual tokens, instruction tokens, proprioceptive state tokens, and action-latent tokens. 
The Semantic Expert processes the $\{\mathbf{X}_V, \mathbf{X}_L\}$, while the Action Expert is dedicated to the kinematic subset $\{\mathbf{X}_S, \mathbf{X}_A\}$.
To support strong cross-modal fusion, the two experts interact at every layer through a shared causal self-attention mechanism. 
Although the hidden sizes of the two streams differ, we use the same number of attention heads and the same per-head dimension in both experts.
At each layer, we compute attention by concatenating the projected Queries, Keys, and Values from the two streams as follows:

\begin{equation}\small\mathbf{T}^{(l)} = \big[\,\mathbf{T}_{\mathrm{sem}}^{(l)} \;\; \mathbf{T}_{\mathrm{act}}^{(l)}\,\big],\quad \mathbf{T}\in\{\mathbf{Q},\mathbf{K},\mathbf{V}\},
\end{equation}
where square brackets $[\,\cdot\,]$ denote concatenation along the sequence dimension. 
This lets the Action Expert attend to the full visual–semantic context directly, injecting high-frequency control information without first compressing it through a discrete text-embedding space.

\noindent\textbf{Continuous Action Chunking via Flow Matching.}
We use flow matching~\cite{lipman2022flow} to model action generation as a conditional distribution over an action chunk $p(\mathbf{a}_{t:t+H} | \mathbf{o}_t)$ where $H$ is a short prediction horizon. 
This formulation naturally produces smooth and potentially action trajectories.
Concretely, the Action Expert learns a time-dependent vector field $v_t(\mathbf{x}, \tau)$ that transports samples from a Gaussian noise distribution $\pi_0(\mathbf{x}) = \mathcal{N}(\mathbf{0}, \mathbf{I})$ to the action-chunk data distribution $\pi_1(\mathbf{x}) \approx \mathbf{a}_{t:t+H}$, conditioned on multimodal context $c$ (Semantic Expert features and proprioception).
During training, given a data sample $\mathbf{x}_1$, a noise sample $\mathbf{x}_0\sim\pi_0$, and a timestep $\tau \sim \mathcal{U}[0,1]$, we form the interpolated point $\mathbf{x}_\tau = (1-\tau)\mathbf{x}_0 + \tau\mathbf{x}_1$ and minimize the flow-matching objective:

\begin{equation}
\mathcal{L}_{\mathrm{FM}} = \mathbb{E}_{\tau,\mathbf{x}_0,\mathbf{x}_1} \Bigl[ \bigl\| v_\theta(\mathbf{x}_\tau, \tau, \mathbf{c}) - (\mathbf{x}_1 - \mathbf{x}_0) \bigr\|^2 \Bigr],
\end{equation}
where $\mathbf{c}$ denotes the multimodal conditioning context combining features from the Semantic Expert with proprioceptive inputs.
At inference time, we integrate the resulting Ordinary Differential Equation (ODE) with an explicit Euler solver to generate the action sequence.

\noindent\textbf{Physically Grounded Unified Action Space.}
To support scalable pre-training across heterogeneous robots---from single-arm grippers to bimanual dexterous hands---we define a physically grounded unified action space $\mathcal{A}_{\text{uni}} \in \mathbb{R}^{D}$. 
This space acts as a superset of all supported physical degrees of freedom (DoF), and is partitioned into semantically aligned subspaces:

\begin{equation}
\mathcal{A}_{\text{uni}} = \mathcal{S}_{\text{eef}} \oplus \mathcal{S}_{\text{joint}} \oplus \mathcal{S}_{\text{gripper}} \oplus \mathcal{S}_{\text{hand}} \oplus \mathcal{S}_{\text{aux}}
\end{equation}
Here, $\mathcal{S}_{\text{eef}}$ represents bimanual end-effector poses (translation and axis–angle rotation) for Cartesian control; 
$\mathcal{S}_{\text{joint}}$ supports direct joint-space commands (e.g., 7-DoF arms); 
$\mathcal{S}_{\text{gripper}}$ encodes parallel-jaw gripper states; 
$\mathcal{S}_{\text{hand}}$ provides higher-dimensional slots for dexterous hands; 
and $\mathcal{S}_{\text{aux}}$ covers auxiliary mechanisms.
To study which action parameterization scales best, we allow flexible action parameterization within $\mathcal{S}_{\text{eef}}$.
Let $\mathbf{T}_{\tau} \in SE(3)$ denote the target end-effector pose at horizon step $\tau$, and $\mathbf{T}_{0}$ be the pose at the start of the action chunk. 
We define a coordinate mapping $\Psi$ with four modes:

\begin{equation}
\Psi(\mathbf{T}_{\tau}) \in
\begin{cases}
    \mathbf{T}_{\tau} \ominus \mathbf{T}_{0} & \text{(World-Rel)} \\
    \mathbf{T}_{\tau} \ominus \mathbf{T}_{\tau-1} & \text{(World-Delta)} \\
    \mathbf{T}_{0}^{-1} \circ \mathbf{T}_{\tau} & \text{(EEF-Rel)} \\
    \mathbf{T}_{\tau-1}^{-1} \circ \mathbf{T}_{\tau} & \text{(EEF-Delta)}
\end{cases}
\end{equation}
where $\ominus$ computes the pose difference in the world frame (translation difference and rotational displacement), and $\circ$ denotes $SE(3)$ composition in the local (end-effector) frame. 
\textit{World/EEF-Rel} express displacement relative to the chunk start, whereas \textit{World/EEF-Delta} capture step-to-step increments.
For each robot embodiment $r$ with native action space $\mathcal{A}_r$, we define an embedding mapping $\phi_r: \mathcal{A}_r \rightarrow \mathcal{A}_{\text{uni}}$ that places robot-specific actions into the appropriate semantic slots of the unified space. 
Dimensions not used by embodiment $r$ are disabled with a binary mask.
This construction encourages the model to learn shared physical priors in subspaces that overlap across robots, while still allowing embodiment-specific control in non-overlapping subspaces.

\begin{algorithm}[h]
\caption{Grouped Blind Ensemble Protocol}
\label{alg:blind_eval}
\begin{algorithmic}[1]
    \Require Model Pool $\mathcal{M}$, Task Set $\mathcal{T}$, Group Size $K$, Trials per model $N$
    \State Divide $\mathcal{M}$ into random non-overlapping groups $\{G_1, \dots, G_M\}$
    \For{each task $\tau \in \mathcal{T}$}
        \For{each model group $G_j$}
            \State $\mathcal{H} \leftarrow \text{Anonymize}(G_j)$ \Comment{Map models to aliases}
            \State $\mathcal{Q} \leftarrow \text{Shuffle}(\{ \underbrace{h_1, \dots, h_1}_{N}, \dots, \underbrace{h_K, \dots, h_K}_{N} \})$ \Comment{Create randomized trial queue}
            \State Initialize results $\mathcal{R}_j \leftarrow \emptyset$
            \While{$\mathcal{Q}$ is not empty}
                \State $h \leftarrow \mathcal{Q}.\text{pop}()$ \Comment{Get next anonymous model}
                \State Operator executes task $\tau$ with policy $\pi_h$
                \State Record outcome $r \in \{0, 1\}$ to $\mathcal{R}_j$
            \EndWhile
            \State \textbf{Break} \Comment{Allow operator rest between groups}
        \EndFor
    \EndFor
    \State \Return Deanonymized aggregate statistics
\end{algorithmic}
\end{algorithm}

\subsection{Grouped Blind Ensemble Evaluation}

Evaluating robotic foundation models is susceptible to experimenter bias.
To enable objective comparisons across training strategies, we propose the Grouped Blind Ensemble Protocol, which implements a double-blind procedure that cleanly separates policy execution (inference) from outcome assessment.
As shown in Algorithm~\ref{alg:blind_eval}, we randomly partition the model pool into manageable groups of size $K$ (typically 4-8). 
For each task, models within the active group are anonymized and evaluated in a randomized order. 
The system dispatches the next policy to run, and the operator functions only as an executor by performing the rollout and recording binary success/failure without access to model identities or versions. 
Grouping serves two purposes: it reduces the influence of human preferences by keeping the operator blind to the tested model, and it supports structured breaks between groups to mitigate fatigue, helping maintain consistent evaluation quality in large-scale studies.



\begin{figure}[h]
\centering
\includegraphics[width=0.7\linewidth]{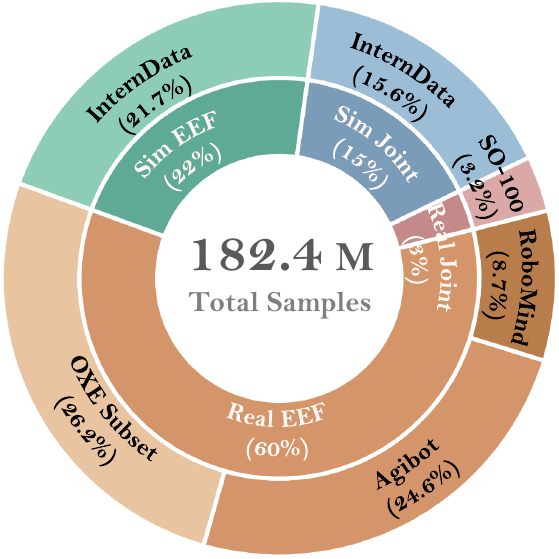}
\caption{
Composition of the balanced pre-training data.
}
\label{fig:data_distribution}
\vspace{-1.5em}
\end{figure}

\section{Pre-training Data and Implementation}
\label{sec:implementation}

In this section, we describe how we construct our heterogeneous corpus for VLA pre-training and summarize the key implementation settings.

\subsection{Large-scale Heterogeneous Robot Data}
Training a generalist VLA requires data that spans diverse embodiments, environments, and control modalities.
We therefore aggregate a large-scale collection of robot manipulation trajectories and organize the sources along two axes: domain (real vs.\ simulation) and control space (end-effector vs.\ joint space). 
This yields four quadrants, which we use to structure and report the composition of our training mixture as follows.

\noindent\textbf{Real-World End-Effector Data.} 
Real-world end-effector (EEF) trajectories form the core of our manipulation pre-training data. 
We include the Open X-Embodiment (OXE) dataset~\cite{o2024openx}, only using its high-quality subsets such as DROID, Bridge, and Fractal.
We further incorporate large-scale proprietary datasets from Agibot~\cite{bu2025agibot}, which cover both gripper-based and dexterous-hand EEF control, and RoboMind~\cite{wu2024robomind,hou2025robomind2}, spanning multiple platforms including Franka, UR5, and AgileX mobile manipulators.

\noindent\textbf{Simulation and Joint-Space Data.} 
To broaden kinematic coverage and improve fine-grained control, we incorporate simulation trajectories from InternData~\cite{chen2025internvla-m1}, including both conventional arm manipulation and dexterous-hand tasks.
To strengthen joint-space modeling, we additionally use data from low-cost teleoperation platforms such as SO-100~\cite{shukor2025smolvla} and real-world humanoid datasets~\cite{tian2025interndata-a1}.

\noindent\textbf{Data Balancing Strategy.} 
A key challenge in co-training is the severe imbalance in dataset scale: raw frame counts range from hundreds of millions to only a few hundred thousand. 
Since most existing VLA training treats each frame transition as an independent sample, naively mixing datasets would cause the largest sources to dominate the gradient. 
We mitigate this issue by applying a dynamic downsampling strategy using a dataset-specific via \texttt{frame\_step\_size}. 
Concretely, we aggressively downsample dense simulation streams and high-frequency real-world logs (e.g., step size 7 for Agibot Gripper; step sizes 4-8 for InternData Sim subsets), while keeping dense sampling for smaller but diverse datasets (e.g., step size 1 for SO-100 and most OXE subsets). 
After balancing, the effective corpus contains approximately 180 million frame transitions. 
Figure~\ref{fig:data_distribution} illustrates the balanced  distribution of the pre-training mixture.

\subsection{Pretraining Implementation Details}

For model configuration, the semantic expert $\mathcal{E}_{\text{sem}}$ is initialized from InternVL-3.5-2B~\cite{wang2025internvl3.5} with hidden size 2048, and the action expert $\mathcal{E}_{\text{act}}$ is a Transformer decoder trained from scratch, comprising 0.7B parameters with hidden size 1024.  
To enable the synchronized attention mechanism, we use a shared per-head dimension in both experts, ensuring that their Query/Key/Value projections can be concatenated despite the mismatch in hidden sizes.
The model operates in the unified action space using relative end-effector (EEF) commands for Cartesian control and absolute values for joint space control.
To assess the effect of regularization in large-scale VLA pre-training, we integrate two widely used stochastic techniques into our training pipeline.

During pretraining, we first mask proprioception by zeroing the input proprioceptive vector with probability $p=0.2$.
Then, we also apply independent dropping of each camera view with $p=0.2$.
Note that at least one view is retained if all are stochastically dropped to ensure valid visual inputs.
We use a two-stage curriculum to investigate whether gradual adaptation outperforms direct joint optimization. 
In Stage 1, we freeze the VLM backbone and optimize only the action expert for 40k steps. 
In Stage 2, we unfreeze the full model and train all parameters for 200k steps. 
All experiments use a global batch size of 256 on 8 NVIDIA A800 GPUs.

\begin{figure}[h]
\centering
\includegraphics[width=0.9\linewidth]{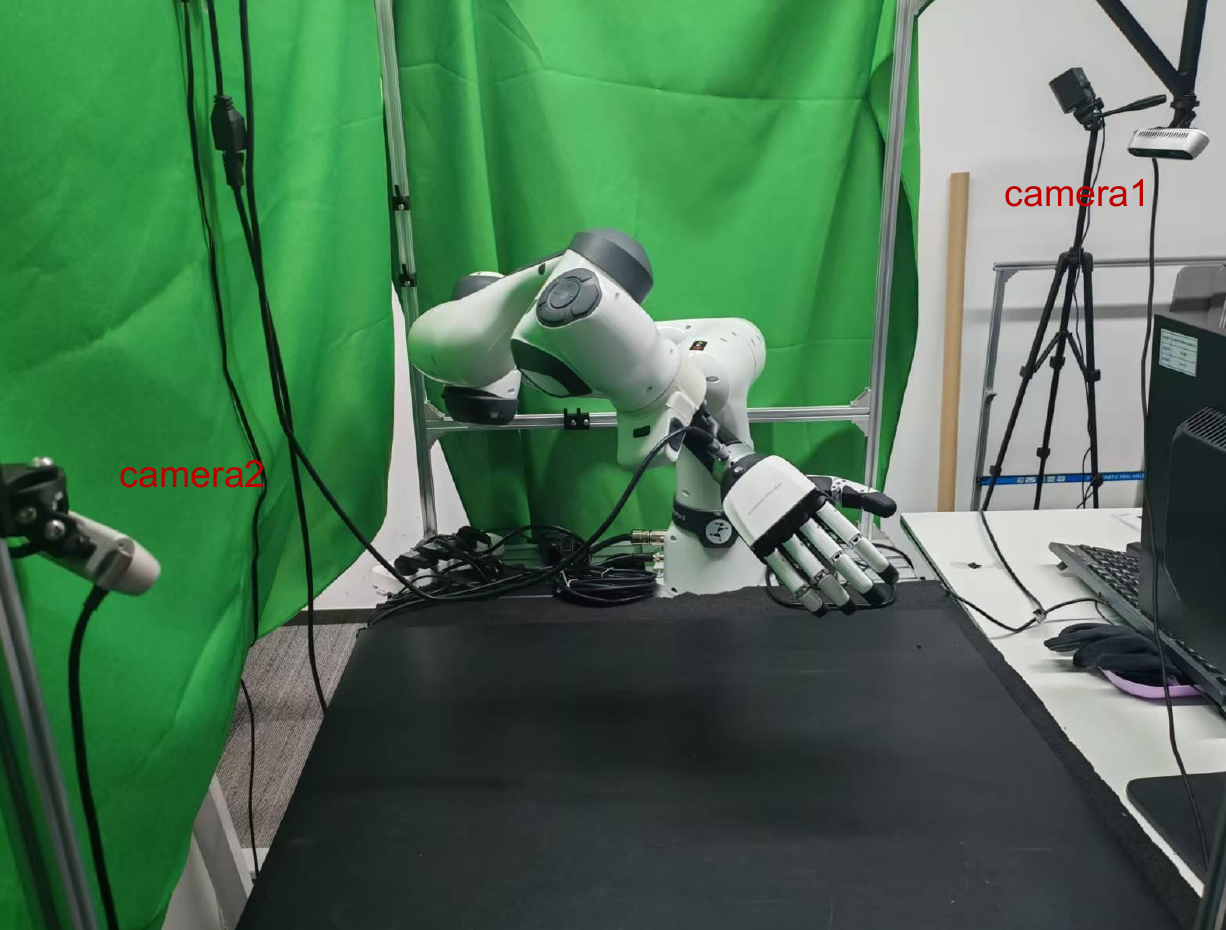}
\caption{Real-world experimental setup.}
\label{fig:real_setup}
\vspace{-1.5em}
\end{figure}

\section{Experiments}
\label{sec:experiments}

In this section, we present a systematic study that disentangles the scaling recipe for VLAs trained on heterogeneous robot data. 
Our analysis centers on three key design dimensions:
\textbf{(1) Physical Alignment} which investigates how coordinate frames and action representations affect robot control precision;
\textbf{(2) Embodiment Mixture} which assess whether aggregating heterogeneous data sources yields positive transfer or introduces interference instead; 
and \textbf{(3) Training Regularization} which evaluates the effectiveness of sensory dropout strategies at scale.

\begin{table*}[!t]
\centering
\small
\caption{Pre-training transfer ablation on Libero 5-shot benchmark under different action space configuration.
}
\label{tab:libero_ablation}
\renewcommand{\arraystretch}{1.25} 
\setlength{\tabcolsep}{3.5pt}

\resizebox{\textwidth}{!}{%
    \begin{tabular}{ll lllll l lllll}
    \toprule
    \multirow{2.5}{*}{\textbf{Action Space}} & \multirow{2.5}{*}{\textbf{Init.}} & \multicolumn{5}{c}{\textbf{Unfrozen VLM}} & & \multicolumn{5}{c}{\textbf{Frozen VLM}} \\
    \cmidrule(lr){3-7} \cmidrule(lr){9-13}
     & & Spatial & Object & Goal & Long & \textbf{Avg} & & Spatial & Object & Goal & Long & \textbf{Avg} \\
    \midrule
    
    \multicolumn{13}{l}{\textit{World-Frame Coordinates}} \\
    \multirow{2}{*}{World-Relative} 
     & Scratch & 90.8 & 93.6 & 78.2 & 74.8 & 84.4 & & 72.2 & 85.0 & 75.4 & 43.6 & 69.1 \\
     & \cg Pretrain 
     & \cg 89.8 \loss{-1.0} & \cg 86.8 \loss{-6.8} & \cg 88.4 \gain{10.2} & \cg 68.8 \loss{-6.0} & \cg 83.5 \loss{-0.9} 
     & & \cg 81.2 \gain{9.0} & \cg 89.8 \gain{4.8} & \cg 75.0 \loss{-0.4} & \cg 52.6 \gain{9.0} & \cg 74.7 \gain{5.6} \\
    \addlinespace[4pt]
    
    \multirow{2}{*}{World-Delta} 
     & Scratch & 90.4 & 93.0 & 82.8 & 73.6 & 85.0 & & 74.2 & 75.0 & 75.0 & 42.6 & 66.7 \\
     & \cg Pretrain 
     & \cg 86.2 \loss{-4.2} & \cg 92.0 \loss{-1.0} & \cg 88.8 \gain{6.0} & \cg 71.0 \loss{-2.6} & \cg 84.5 \loss{-0.5} 
     & & \cg 82.6 \gain{8.4} & \cg 89.6 \gain{14.6} & \cg 73.8 \loss{-1.2} & \cg 53.2 \gain{10.6} & \cg 74.8 \gain{8.1} \\
    \midrule
    
    \multicolumn{13}{l}{\textit{End-Effector Coordinates}} \\
    \multirow{2}{*}{EEF-Relative} 
     & Scratch & 87.6 & 93.0 & 76.4 & 70.6 & 81.9 & & 73.0 & 84.2 & 70.6 & 39.6 & 66.9 \\
     & \cg Pretrain 
     & \cg 86.0 \loss{-1.6} & \cg 90.0 \loss{-3.0} & \cg 88.6 \gain{12.2} & \cg 73.4 \gain{2.8} & \cg 84.5 \gain{2.6} 
     & & \cg 79.8 \gain{6.8} & \cg 90.6 \gain{6.4} & \cg 76.6 \gain{6.0} & \cg 53.2 \gain{13.6} & \cg 75.1 \gain{8.2} \\
    \addlinespace[4pt]
    
    \multirow{2}{*}{EEF-Delta} 
     & Scratch & 87.6 & 93.0 & 72.0 & 67.0 & 79.9 & & 73.2 & 79.8 & 70.4 & 41.0 & 66.1 \\
     & \cg Pretrain 
     & \cg 85.6 \loss{-2.0} & \cg 89.8 \loss{-3.2} & \cg 87.4 \gain{15.4} & \cg 66.2 \loss{-0.8} & \cg 82.3 \gain{2.4} 
     & & \cg 77.8 \gain{4.6} & \cg 88.2 \gain{8.4} & \cg 73.6 \gain{3.2} & \cg 48.4 \gain{7.4} & \cg 72.0 \gain{5.9} \\
    \bottomrule
    \end{tabular}%
}
\end{table*}

Our evaluation is designed to rigorously stress-test the quality of the pre-trained representations in both standardized simulation benchmarks and real-world deployments.

\noindent\textbf{Simulation Benchmarks.}
We first evaluate downstream transfer on LIBERO~\cite{liu2023libero} and RoboCasa~\cite{nasiriany2024robocasa}. 
In both benchmarks, we follow a multi-task fine-tuning protocol in which a single unified policy is trained jointly across all tasks.
For LIBERO, we fine-tune on the union of all four task suites (Spatial, Object, Goal, Long) under a low-data regime of 5 demonstrations per task (5-shot).
For RoboCasa, we fine-tune on all 24 kitchen tasks using the standard few-shot setup of 50 human demonstrations per task.

\begin{table}[h]
\centering
\resizebox{\columnwidth}{!}{%
    \begin{tabular}{ll llll}
    \toprule
    \multirow{2.5}{*}{\textbf{Action Space}} & \multirow{2.5}{*}{\textbf{Init.}} & Pick/ & Doors/ & \multirow{2.5}{*}{Other} & \multirow{2.5}{*}{\textbf{Avg}} \\
     & & Place & Drawers & & \\
    \midrule
    
    \multicolumn{6}{l}{\textit{World-Frame Coordinates}} \\
    \multirow{2}{*}{World-Relative} & Scratch & 20.0 & 60.0 & 40.0 & 38.3 \\
     & \cg Pretrain 
     & \cg 22.5 \gain{2.5} & \cg 58.3 \loss{-1.7} & \cg 45.0 \gain{5.0} & \cg 40.8 \gain{2.5} \\
    \addlinespace[4pt]
    \multirow{2}{*}{World-Delta} & Scratch & 21.0 & 62.3 & 46.0 & 41.8 \\
     & \cg Pretrain 
     & \cg 23.8 \gain{2.8} & \cg 60.0 \loss{-2.3} & \cg 45.0 \loss{-1.0} & \cg 41.7 \loss{-0.1} \\
    \midrule
    
    \multicolumn{6}{l}{\textit{End-Effector Coordinates}} \\
    \multirow{2}{*}{EEF-Relative} & Scratch & 33.0 & 57.3 & 47.4 & 45.1 \\
     & \cg Pretrain 
     & \cg 35.3 \gain{2.3} & \cg 62.3 \gain{5.0} & \cg 54.4 \gain{7.0} & \cg 50.0 \gain{4.9} \\
    \addlinespace[4pt]
    
    \multirow{2}{*}{EEF-Delta} & Scratch & 26.0 & 63.7 & 44.8 & 43.3 \\
     & \cg Pretrain 
     & \cg 36.3 \gain{10.3} & \cg 56.7 \loss{-7.0} & \cg 49.0 \gain{4.2} & \cg 46.7 \gain{3.4} \\
    \bottomrule
    \end{tabular}%
}
\caption{Pre-training transfer ablation on RoboCasa benchmark under different action space configuration.}
\label{tab:robocasa_ablation}
\end{table}

\noindent\textbf{Real-Robot Protocol.}
As shown in Figure~\ref{fig:real_setup}, our experimental setup features a Franka Panda arm equipped with two RGB cameras positioned on the left and right sides. 
We evaluate four tasks that probe complementary capabilities.
\textit{Stack Bowls} focuses on precision with a maximum score of 3; 
the task requires placing three bowls into a plate, and we award 1 point for each successful placement.
\textit{Pick-to-Drawer} examines long-horizon planning with a maximum score of 5; 
the score comprises 1 point for opening the drawer, 1 point for each of the three objects placed inside, and 1 point for closing the drawer.
\textit{Wipe Board} involves dynamic motion with a maximum score of 4; 
we assign 1 point for each specific step: 
picking up the cloth, initiating the wiping motion, wiping the surface clean, and placing the cloth back.
\textit{Water Plant} targets non-rigid object interaction with a maximum score of 3; 
the sequence includes picking up the spray can, aiming at the plant, and pressing the trigger, where each action contributes 1 point.
For each task, we conduct 10 trials, calculate the total score, and multiply it by a fixed coefficient to obtain a percentage.
During inference, we adopt a grouped blind ensemble protocol: 
for each experimental setting, the operator remains unaware of the model variant to minimize subjective bias.

\begin{figure*}[h]
\centering
\includegraphics[width=.8\linewidth]{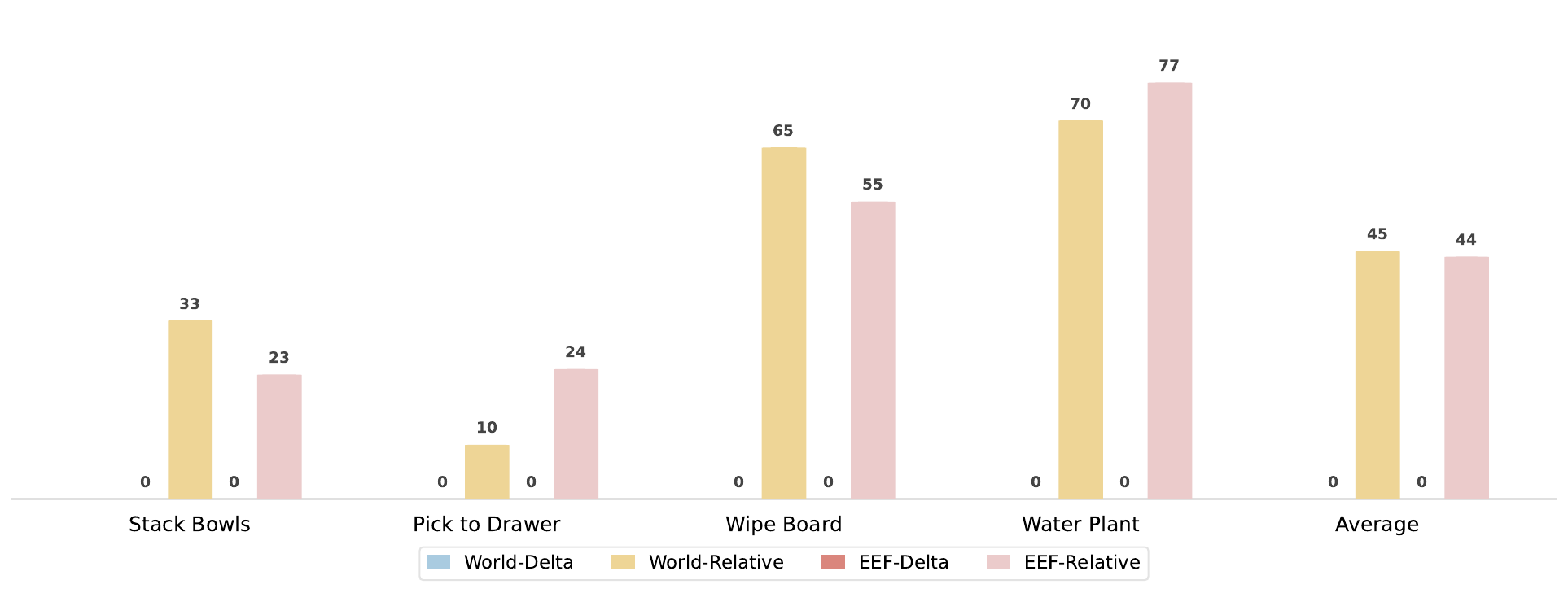}
\caption{
Real-world blind evaluation of different action spaces.
}
\label{fig:real_action_bar}
\end{figure*}

\subsection{Exploration of Physical Alignment}
\label{sec:exp_action}
A key challenge in leveraging heterogeneous robot data is to standardize the configuration of action representation for robots with different kinematics and base placements.
To study this, we evaluate four coordinate parameterizations: \textbf{World-Relative}, \textbf{World-Delta}, \textbf{EEF-Relative}, and \textbf{EEF-Delta}, and compare policies trained from scratch with those fine-tuned from our heterogeneous pre-training corpus.

\noindent\textbf{Simulation Results.}
Table~\ref{tab:libero_ablation} shows that the coordinate frame strongly influences the performance of pre-training transfer.
On LIBERO, scratch-trained policies paradoxically perform better with world-frame actions compared to using EEF coordinates, likely because fixed camera extrinsics and bounded workspace make absolute positions easy to exploit.
In contrast, pre-training yields negative transfer for world-frame variants (-0.9\%, -0.5\%) but consistent gains for EEF-frame variants (+2.6\%, +2.4\%). 
This inversion indicates that world coordinates can overfit single-environment regularities, whereas EEF coordinates generalize better across varying cameras and robot bases in large-scale data.

\begin{table*}[t]
\centering
\small
\caption{
LIBERO 5-shot transfer performance across incremental pre-training mixtures.
}
\label{tab:data_libero}
\renewcommand{\arraystretch}{1.25}
\setlength{\tabcolsep}{1.5pt} 

\resizebox{\textwidth}{!}{%
\begin{tabular}{l ccccc c ccccc}
\toprule
\multirow{2.5}{*}{\textbf{Pre-train Mixture}} & \multicolumn{5}{c}{\textbf{Frozen VLM}} & & \multicolumn{5}{c}{\textbf{Unfrozen VLM}} \\
\cmidrule(lr){2-6} \cmidrule(lr){8-12}
 & Spat & Obj & Goal & Long & \textbf{Avg} & & Spat & Obj & Goal & Long & \textbf{Avg} \\
\midrule

Scratch 
 & 73.0 & 84.2 & 70.6 & 39.6 & 66.9 
 & & 87.6 & 93.0 & 76.4 & 70.6 & 81.9 \\
\midrule

$\mathcal{D}_1$: OXE Only
 & \textbf{87.6} \gain{14.6} & 88.2 \gain{4.0} & 78.8 \gain{8.2} & \textbf{54.6} \gain{15.0} & \textbf{77.3} \gain{10.4}
 & & \textbf{89.2} \gain{1.6} & 89.2 \loss{-3.8} & \textbf{92.8} \gain{16.4} & \textbf{74.4} \gain{3.8} & \textbf{86.4} \gain{4.5} \\

\hspace{0.5em}$\llcorner$ $\mathcal{D}_2$: $\mathcal{D}_1$ + Real EEF
 & 84.4 \loss{-3.2} & 86.0 \loss{-2.2} & 77.4 \loss{-1.4} & 47.4 \loss{-7.2} & 73.8 \loss{-3.5}
 & & 86.2 \loss{-3.0} & \textbf{94.6} \gain{5.4} & 88.8 \loss{-4.0} & 66.4 \loss{-8.0} & 84.0 \loss{-2.4} \\

\hspace{1.0em}$\llcorner$ $\mathcal{D}_3$: $\mathcal{D}_2$ + Sim EEF
 & 76.0 \loss{-8.4} & 82.2 \loss{-3.8} & \textbf{79.0} \gain{1.6} & 51.2 \gain{3.8} & 72.1 \loss{-1.7}
 & & 87.4 \gain{1.2} & 91.4 \loss{-3.2} & 86.2 \loss{-2.6} & 69.6 \gain{3.2} & 83.7 \loss{-0.3} \\

\hspace{1.5em}$\llcorner$ $\mathcal{D}_4$: $\mathcal{D}_3$ + Joint
 & 79.8 \gain{3.8} & \textbf{90.6} \gain{8.4} & 76.6 \loss{-2.4} & 53.2 \gain{2.0} & 75.1 \gain{3.0}
 & & 86.0 \loss{-1.4} & 90.0 \loss{-1.4} & 88.6 \gain{2.4} & 73.4 \gain{3.8} & 84.5 \gain{0.8} \\

\bottomrule
\end{tabular}%
} 
\end{table*}

\begin{table}[!t]
\centering
\resizebox{\columnwidth}{!}{%
    \begin{tabular}{l cccc}
    \toprule
    \multirow{2.5}{*}{\textbf{Pre-train Mixture}} & Pick/ & Doors/ & \multirow{2.5}{*}{Other} & \multirow{2.5}{*}{\textbf{Avg}} \\
     & Place & Drawers & & \\
    \midrule
    
    Scratch
     & 33.0 & 57.3 & 47.4 & 45.1 \\
    \midrule
    
    $\mathcal{D}_1$: OXE Only
     & \textbf{42.0} \gain{9.0} & \textbf{72.3} \gain{15.0} & 54.2 \gain{6.8} & \textbf{54.7} \gain{9.6} \\
    
    \hspace{0.5em}$\llcorner$ $\mathcal{D}_2$: $\mathcal{D}_1$ + Real EEF
     & 38.5 \loss{-3.5} & 60.5 \loss{-11.8} & 47.4 \loss{-6.8} & 48.8 \loss{-5.9} \\
    
    \hspace{1.0em}$\llcorner$ $\mathcal{D}_3$: $\mathcal{D}_2$ + Sim EEF
     & 36.3 \loss{-2.2} & 56.7 \loss{-3.8} & \textbf{56.0} \gain{8.6} & 49.6 \gain{0.8} \\
    
    \hspace{1.5em}$\llcorner$ $\mathcal{D}_4$: $\mathcal{D}_3$ + Joint
     & 35.3 \loss{-1.0} & 62.3 \gain{5.6} & 54.4 \loss{-1.6} & 50.0 \gain{0.4} \\
    \bottomrule
    \end{tabular}%
}
\caption{
RoboCasa 50-shot transfer performance across incremental pre-training mixtures.
}
\label{tab:data_robocasa}
\end{table}

The benefits of pre-training become even more pronounced when the VLM backbone is frozen.
In this regime, the VLM backbone remains fixed during fine-tuning, the performance relies heavily on the high-quality representations initialized from pre-training, including both the aligned VLM and the Action Expert.
We observe substantial gains over the scratch baseline across all action spaces.
Most notably, EEF-Relative achieves the highest average success rate of 75.1\% with the largest improvement of +8.2\%.
This confirms that the end-effector space is the most effective choice for preserving and transferring physical priors when the visual backbone is static.

These trends are even more pronounced on the geometrically diverse RoboCasa benchmark (Table~\ref{tab:robocasa_ablation}).
The EEF-Relative representation exhibits the most reliable scaling behavior, increasing average success from 45.1\% to 50.0\%.
Gains are consistent across task categories, including a +5.0\% improvement on articulated-object tasks (Doors/Drawers) and a +7.0\% increase on the remaining tasks. 
In contrast, alternative action representations are unstable or ineffective.
For instance, World-Delta shows essentially no benefit and even slight negative transfer (-0.1\%).
while EEF-Delta degrades substantially on Doors/Drawers (-7.0\%) despite pretraining

\noindent\textbf{Real-World Results.}
Beyond simulation, we further corroborate these findings with blinded real-world evaluations (Figure~\ref{fig:real_action_bar}).
We note that all Delta actions exhibit jittering in place on the real robot, rendering them unable to execute tasks and resulting in a 0\% success rate. 
In contrast, Relative actions execute effectively. 
We observe no significant performance difference between World-Relative and EEF-Relative in real-world settings.

\begin{figure*}[h!]
\centering
\includegraphics[width=.8\linewidth]{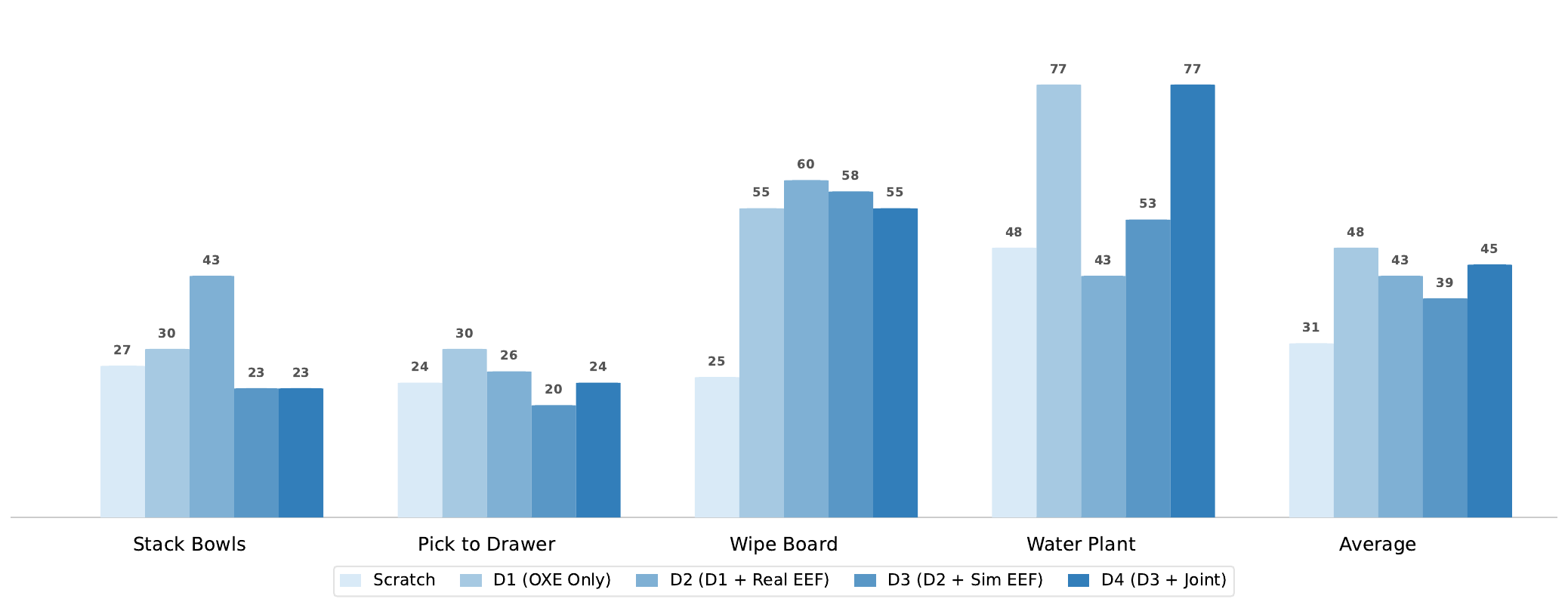}
\caption{
Real-world blind evaluation of different pre-training data mixtures.
}
\label{fig:real_data_bar}
\vspace{-1em}
\end{figure*}

\subsection{Exploration of Embodiment Mixture}
\label{sec:exp_data_scale}
With the action space fixed to EEF-Relative, we study how performance changes as we scale heterogeneous pre-training data using a cumulative inclusion protocol.
Specifically, we form four progressively richer data mixtures to isolate the marginal contribution of each data source:
\textbf{D1 (OXE Only)} utilizes standard public datasets.
\textbf{D2 (+Real EEF)} adds real-world end-effector trajectories from additional robot embodiments.
\textbf{D3 (+Sim EEF)} further includes simulation end-effector data.
Finally, \textbf{D4 (+Joint)} additionally incorporates joint-space demonstrations, projected into the unified model.

In contrast to language-model scaling behavior, enlarging the robotic pre-training corpus does \textbf{NOT} yield monotonic gains.
As shown in Table \ref{tab:data_libero}, $\mathcal{D}_1$ provides the strongest pre-training transfer on LIBERO, reaching 77.3\% average success in the Frozen VLM setting (+10.4\% over training from scratch).
However, simply increasing data diversity degrades performance.
The addition of heterogeneous real-robot EEF data ($\mathcal{D}_2$) drops success to 73.8\%, and including simulated EEF trajectories ($\mathcal{D}_3$) further reduces it to 72.1\%.
Projecting joint-space data in $\mathcal{D}_4$ recovers part of the loss (+3.0\%), but still does not surpass the OXE-only baseline.
This negative trend is even more pronounced on RoboCasa (Table~\ref{tab:data_robocasa}).
$\mathcal{D}_1$ sets a high-water mark of 54.7\%, while adding diverse real-world EEF data in $\mathcal{D}_2$ immediately decreases performance to 48.8\%. 
Later mixtures ($\mathcal{D}_3, \mathcal{D}_4$) offer only marginal recovery, leaving the final model nearly \textbf{5\%} below $\mathcal{D}_1$ overall and \textbf{10.0\%} worse on articulated-object tasks (Doors/Drawers).
Collectively, these results suggests thatnaively pooling structurally disparate robot datasets induces destructive interference rather than improved transfer.

\noindent\textbf{Real-World Results.}
As shown in Figure \ref{fig:real_data_bar}.
Real-world experimental results demonstrate that pre-training models offer a significant advantage, as all pre-training variants ($D_1$ through $D_4$) substantially outperform the training-from-scratch (Scratch) baseline. 
Specifically, while the $D_1$ scheme using only the OXE dataset maintains robust performance across multiple tasks, the introduction of simulated end-effector (Sim EEF) trajectories in the $D_3$ stage leads to varying degrees of performance degradation in tasks such as ``Stack Bowls," ``Pick to Drawer," and ``Wipe Board," further validating the potential destructive interference caused by heterogeneous data.

\begin{table}[t]
\centering
\small
\caption{
Impact of training regularization and scheduling on LIBERO 5-shot performance.
}
\label{tab:ablation_dynamics}
\renewcommand{\arraystretch}{1.2} 
\setlength{\tabcolsep}{4.5pt}     

    \begin{tabular}{cc c ccccc}
    \toprule
    \multicolumn{2}{c}{\textbf{Mask Ratio}} & \multirow{2.5}{*}{\textbf{Schedule}} & \multicolumn{5}{c}{\textbf{Success Rate (\%)}} \\
    \cmidrule(lr){1-2} \cmidrule(lr){4-8}
    State & View & & Spat & Obj & Goal & Long & \textbf{Avg} \\
    \midrule
    
    \rowcolor{highlightrow}
    0.2 & 0.2 & 2-Stage & 86.0 & 90.0 & 88.6 & 73.4 & 84.5 \\
    \midrule
    
    0 & 0.2 & 2-Stage & 88.2 & \textbf{96.0} & 87.6 & 69.0 & 85.2 \\
    0.5 & 0.2 & 2-Stage & \textbf{88.8} & 90.4 & 86.8 & 70.0 & 84.0 \\
    \addlinespace[2pt]
    
    0.2 & 0 & 2-Stage & 87.0 & 92.2 & \textbf{89.2} & \textbf{74.0} & 85.6 \\
    0.2 & 0.5 & 2-Stage & 85.8 & 93.2 & 86.6 & 72.2 & 84.5 \\
    \midrule
    
    0.2 & 0.2 & Stage 2 Only & 88.2 & 93.8 & 87.0 & \textbf{74.0} & \textbf{85.8} \\
    
    \bottomrule
    \end{tabular}%
\end{table}

\subsection{Exploration of Training Regularization}
\label{sec:exp_dynamics}

We evaluate how sensory dropout ($p_{\text{state}}, p_{\text{view}}$) and staged training curricula affect model generalization.
Results in Table~\ref{tab:ablation_dynamics} call into question the need for these widely used regularization practices at scale.

\noindent\textbf{Limited Benefit from Explicit Regularization.}
Contrary to common assumptions, stochastic modality dropout does not provide a consistent improvement.
Disabling visual dropout ($p_{\text{view}}=0$) increases success to 85.6\% versus the balanced baseline (84.5\%), whereas heavy proprioceptive masking reduces performance.
This pattern may suggest that the natural diversity of the pre-training corpus already serves as an effective regularizer, making additional noise injection unnecessary, or even harmful.

\noindent\textbf{Direct Optimization is Sufficient}
The two-stage alignment curriculum is similarly non-essential.
Directly fine-tuning the full model end-to-end (``Stage 2 Only'') attains the best average success 85.8\%, outperforming the multi-stage schedule.
This indicates that the initialized action expert can quickly co-adapt with the VLM backbone under joint training, supporting a simpler and more efficient optimization pipeline.
\begin{table*}[t]
\centering
\small
\caption{
Benchmarking against representative generalist policies on LIBERO and RoboCasa.
}
\label{tab:sota_main}
\renewcommand{\arraystretch}{1.25} 
\setlength{\tabcolsep}{4pt}        

    \begin{tabular}{l ccccc c cccc}
    \toprule
    \multirow{2.5}{*}{\textbf{Model}} & \multicolumn{5}{c}{\textbf{LIBERO}} & & \multicolumn{4}{c}{\textbf{RoboCasa}} \\
    \cmidrule(lr){2-6} \cmidrule(lr){8-11}
     & Spatial & Object & Goal & Long & \textbf{Avg} & & Pick/Place & Doors/Drawers & Other & \textbf{Avg} \\
    \midrule
    
    GR00T-N1 
     & 94.4 & 97.6 & 93.0 & 90.6 & 93.9 
     & & 18.6 & 50.2 & 39.1 & 36.0 \\
     
    $\pi_0$ 
     & 98.0 & 96.8 & 94.4 & 88.4 & 94.4 
     & & 14.0 & 53.1 & \textbf{58.5} & 42.4 \\
     
    $\pi_{0.5}$ 
     & \textbf{98.8} & \textbf{98.2} & 98.0 & 92.4 & 96.9 
     & & 21.5 & 57.8 & 44.9 & 41.4 \\
     
    \midrule
    
    \rowcolor{highlightrow} 
    \textbf{Ours (Base)} 
     & 98.4 & 96.2 & \textbf{98.4} & \textbf{98.2} & \textbf{97.9} 
     & & \textbf{35.3} & \textbf{62.3} & 54.4 & \textbf{50.0} \\
     
    \bottomrule
    \end{tabular}%
\end{table*}

\begin{figure*}[t]
\centering
\includegraphics[width=.76\textwidth]{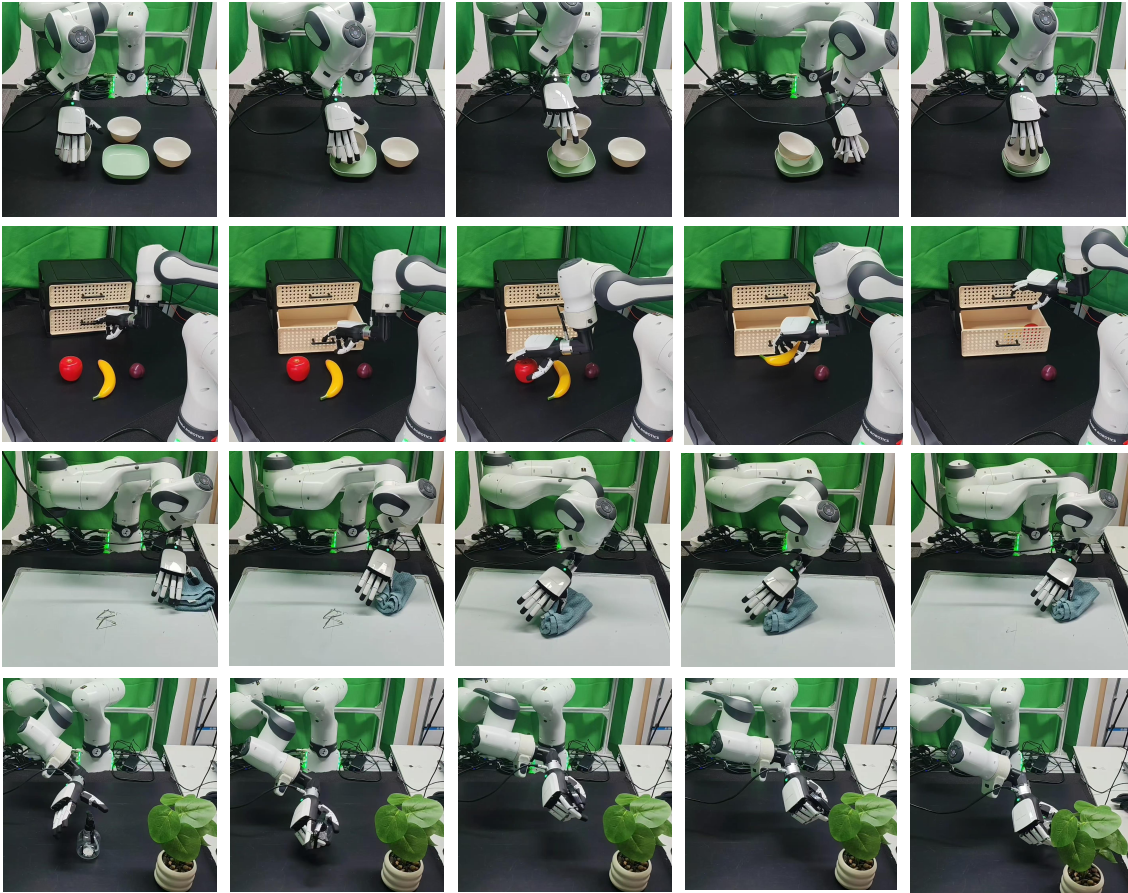}
\caption{
Qualitative rollout sequences across diverse real-world tasks.
}
\label{fig:rollouts}
\vspace{-1em}
\end{figure*}

\subsection{Comparison with Representative Generalist Policies.}
\label{sec:exp_sota}

To validate our testbed, we benchmark our standard implementation against against representative generalist VLA policies ($\pi_0$~\cite{black2024pi_0}, $\pi_{0.5}$~\cite{intelligence2025pi05} and $\text{GR00T-N1}$~\cite{nvidia2025gr00t}) under a matched 50-shot fine-tuning protocol.
As reported in Table~\ref{tab:sota_main}, our base model is competitive without task-specific tuning or specialized optimizations, reaching 97.9\% average success on LIBERO and 50.0\% on RoboCasa benchmarks.
These results indicate that our platform provides a strong and representative foundation, so our experimental conclusions reflect behavior in a high-performance regime aligned with current advances in robot learning.

\subsection{Qualitative Analysis}

Figure~\ref{fig:rollouts} presents rollout sequences for four real-world tasks in our blind evaluation protocol.
The first row shows the ``Stack Bowls" task which focuses on precision, and the challenge lies in the precise consecutive grasping and stacking of three bowls. 
The second row is the ``Pick-to-Drawer" task which requires long-horizon planning to complete multiple subtasks (opening, placing, closing), and it is easy to fail in intermediate stages.
The third row is the ``Wipe Board" task involving dynamic motion and surface contact, which is prone to incomplete wiping. 
The fourth row is the ``Water Plant" task targeting fine-grained tool manipulation, where the main difficulties are knocking over the bottle and failing to press the trigger accurately.

\section{Conclusion}
\label{sec:conclusion}

In this work, we present a systematic study on scaling Vision-Language-Action (VLA) models with heterogeneous robot data. 
To ensure reliable evaluation, we introduce a Grouped Blind Ensemble protocol that minimizes human bias in real-world experiments. 
Our results show that simply increasing data scale does not automatically lead to better performance. 
First, the EEF-Relative action space proves to be the most effective choice for handling diverse robot kinematics. 
Second, scaling across different robots is difficult; simply mixing diverse data often reduces performance, indicating that careful data alignment is essential. 
Finally, complex training techniques like sensory dropout do not consistently bring improvements, suggesting that simple training recipes are often sufficient. 
Overall, this study offers practical guidance for effectively training general-purpose VLA models.

\bibliographystyle{plainnat}
\bibliography{references}

\newpage
\clearpage
\appendix



\subsection{Data Mixture Statistics}
\label{app:data}

Table~\ref{tab:data_stats} details the composition of the balanced pre-training corpus. We calculate the effective frame count $N_{\text{eff}}$ using the raw frame count $N_{\text{raw}}$ and the sampling step size $S$:
\begin{equation}
    N_{\text{eff}} = \lfloor N_{\text{raw}} / S \rfloor
\end{equation}

The final balanced dataset contains approximately 182.4 million frames. We assign larger step sizes ($S \in \{3,4, 7, 8\}$) to the massive Agibot and InternData datasets. This keeps them balanced with the Open X-Embodiment (OXE) subsets.

\begin{table}[h]
\centering
\caption{Detailed statistics of the data mixture.}
\label{tab:data_stats}
\resizebox{\columnwidth}{!}{
\begin{tabular}{l|l|c|r|r}
\toprule
\textbf{Category} & \textbf{Dataset Source} & \textbf{$S$} & \textbf{$N_{\text{raw}}$} & \textbf{$N_{\text{eff}}$} \\
\midrule
\multirow{15}{*}{\textbf{Real EEF}} 
 & OXE - DROID & 1 & 27.04M & 27.04M \\
 & OXE - Bridge & 1 & 1.89M & 1.89M \\
 & OXE - BC-Z & 1 & 5.47M & 5.47M \\
 & OXE - Language Table & 1 & 7.05M & 7.05M \\
 & OXE - Fractal & 1 & 3.79M & 3.79M \\
 & OXE - Kuka & 1 & 2.46M & 2.46M \\
 & \textit{Subtotal (OXE)} & & \textit{47.70M} & \textbf{47.70M} \\
 \cmidrule{2-5}
 & Agibot Gripper & 7 & 249.04M & 35.58M \\
 & Agibot Dexterous & 1 & 9.29M & 9.29M \\
 & RoboMind AgileX & 1 & 5.10M & 5.10M \\
 & RoboMind UR & 1 & 1.34M & 1.34M \\
 & RoboMind Franka & 1 & 3.34M & 3.34M \\
 & RoboMind Tienkung (Gello) & 1 & 2.99M & 2.99M \\
 & RoboMind Tienkung (Xsens) & 1 & 3.18M & 3.18M \\
 \cmidrule{2-5}
 & \textit{Subtotal} & & \textit{321.98M} & \textbf{108.52M} \\
\midrule
\multirow{3}{*}{\textbf{Sim EEF}} 
 & InternData-M1 (Franka) & 4 & 93.39M & 23.35M \\
 & InternData-A1 (Franka) & 3 & 48.82M & 16.27M \\
 \cmidrule{2-5}
 & \textit{Subtotal} & & \textit{142.21M} & \textbf{39.62M} \\
\midrule
\multirow{4}{*}{\textbf{Sim Joint}} 
 & InternData-A1 (Lift2) & 8 & 92.29M & 11.54M \\
 & InternData-A1 (Split-Aloha) & 8 & 90.54M & 11.31M \\
 & InternData-A1 (Genie) & 1 & 5.56M & 5.56M \\
 \cmidrule{2-5}
 & \textit{Subtotal} & & \textit{188.39M} & \textbf{28.41M} \\
\midrule
\multirow{3}{*}{\textbf{Real Joint}} 
 & SO-100 & 1 & 5.02M & 5.02M \\
 & InternData-A1 (Real Genie) & 1 & 0.92M & 0.92M \\
 \cmidrule{2-5}
 & \textit{Subtotal} & & \textit{5.94M} & \textbf{5.94M} \\
\midrule
\textbf{Total} & & & \textit{658.52M} & \textbf{182.49M} \\
\bottomrule
\end{tabular}
}
\end{table}

\subsection{Training Hyperparameters}
\label{app:training_recipe}

We conduct pre-training in two stages. 
In the first stage, we freeze the VLM backbone and train the action expert for 40,000 steps with a learning rate of $1 \times 10^{-4}$. 
In the second stage, we train the full model for 200,000 steps with a learning rate of $2 \times 10^{-5}$. 
Throughout pre-training, we use a global batch size of 256. 
We use the AdamW optimizer with a cosine learning rate scheduler, a warmup ratio of 0.05, and a weight decay of $1 \times 10^{-5}$.

For downstream fine-tuning, we update the full model with a global batch size of 128 and a learning rate of $1 \times 10^{-4}$. 
We train for 30,000 steps on the LIBERO 5-shot benchmark. 
For the full LIBERO benchmark and RoboCasa, we train for 60,000 steps.

\subsection{Real-World Task Details}
\label{app:real_eval}

We evaluate the policies on a Franka Panda robot with two RGB cameras. Each task consists of 10 evaluation trials with randomized object initialization. 
We record success using the following multi-stage criteria.

\begin{itemize}[leftmargin=*]
    \item \textbf{Stack Bowls (Max Score: 3)} \\
    Three bowls are initialized at random positions on the tabletop. The robot must identify, grasp, and stack them into a target plate sequentially. 
    \begin{itemize}
        \item \textbf{Point 1:} Successfully grasp the first bowl and place it into the plate.
        \item \textbf{Point 2:} Grasp the second bowl and stack it inside the first one.
        \item \textbf{Point 3:} Grasp the third bowl and stack it inside the second one.
    \end{itemize}
    \textit{Difficulty:} The main difficulty is to precisely grasp and stack three bowls in a row.

    \item \textbf{Pick-to-Drawer (Max Score: 5)} \\
    This is a long-horizon task involving articulated object manipulation. The scene contains a closed drawer and three distinct objects.
    \begin{itemize}
        \item \textbf{Point 1:} Grasp the handle and fully open the drawer.
        \item \textbf{Point 2:} Pick up Object A and place it inside the drawer.
        \item \textbf{Point 3:} Pick up Object B and place it inside the drawer.
        \item \textbf{Point 4:} Pick up Object C and place it inside the drawer.
        \item \textbf{Point 5:} Push the drawer back to a fully closed state.
    \end{itemize}
    \textit{Difficulty:} Since this task involves many steps, failures often occur during the intermediate stages.

    \item \textbf{Wipe Board (Max Score: 4)} \\
    This task evaluates dynamic contact-rich manipulation. A sponge is placed on the table, and the goal is to wipe a whiteboard surface.
    \begin{itemize}
        \item \textbf{Point 1:} Successfully grasp the sponge from the table.
        \item \textbf{Point 2:} Establish contact between the sponge and the board.
        \item \textbf{Point 3:} Execute a continuous wiping motion.
        \item \textbf{Point 4:} Return the sponge to a designated area.
    \end{itemize}
    \textit{Difficulty:} A common failure is that the robot does not wipe the surface completely clean.
    
    \item \textbf{Water Plant (Max Score: 3)} \\
    The robot interacts with a spray bottle to water a plant. This requires grasping an object with a specific functional orientation.
    \begin{itemize}
        \item \textbf{Point 1:} Grasp the spray bottle by the handle.
        \item \textbf{Point 2:} Reorient the bottle so the nozzle aims at the plant.
        \item \textbf{Point 3:} Press the trigger mechanism.
    \end{itemize}
    \textit{Difficulty:} Common failures include knocking over the bottle or failing to press the trigger accurately.
\end{itemize}




\end{document}